\documentclass{article}

 \usepackage[final]{neurips_2025}

\usepackage[utf8]{inputenc} %
\usepackage[T1]{fontenc}    %

\usepackage{xurl}
\usepackage{hyperref}       %
\usepackage{booktabs}       %
\usepackage{amsfonts}       %
\usepackage{nicefrac}       %
\usepackage{microtype}      %
\usepackage{xcolor}         %
\usepackage{graphicx}
\usepackage{multirow}

\title{It's complicated: the relationship of algorithmic fairness and the non‑discrimination provisions for high‑risk systems in the EU AI Act}

\author{%
  Kristof Meding\\
  University Tübingen\\
  CZS Institute for AI and Law \\
Germany \\
  \texttt{kristof.meding@uni-tuebingen.de} \\
}

\begin{document}

\maketitle

\begin{abstract}
What constitutes a fair decision? This question is not only difficult to answer for humans but becomes more challenging when Artificial Intelligence (AI) models are used. In light of problematic algorithmic outcomes, the EU has recently passed the AI Act, which mandates specific rules for (among others) high-risk systems. For high-risk systems, the AI Act might point towards traditional legal non-discrimination regulations and machine learning based algorithmic fairness concepts. This paper aims to bridge these two concepts as reflected in the AI Act by providing: (1) a high-level introduction of the AI Act targeting computer science-oriented scholars, and (2) an analysis of the relationship between the AI Act’s legal non-discrimination and algorithmic fairness provisions. Finally, we consider future steps for the application of non-discrimination law and the AI Act’s provisions. This paper serves as a foundation for future interdisciplinary collaboration between legal scholars and machine learning researchers with a computer science background by reducing the gap between these two fields. 
\end{abstract}

\section{Introduction}
How can we ensure fair Artificial Intelligence (AI) systems? When regulating the digital world, particularly AI systems, this question becomes fundamental. While the question is also crucial in a non-digital world, it becomes increasingly pressing in the digital world. The European Union (EU) has become one of the forerunners in regulating the digital age. To improve the EU market and to promote the uptake of human-centric and trustworthy AI, the European Union (EU) recently adopted the Artificial Intelligence Act (AIA)\footnote{Regulation (EU) 2024/1689.}, see Article 1 AIA. 
While the AIA is partly a product safety regulation \citep{kaminski2025american,almada2023eu} --- specifically, see Chapter III AIA --- it also incorporates aspects of ``ethical principles'' \citep{nolte2024robustness} within and outside the EU (see recital 27 AIA).

Fair algorithmic processing also matters from a computational perspective. Over the past decade, algorithmic fairness has become a well-established field within machine learning that focuses on defining, mitigating, and evaluating the discriminatory behaviour of Artificial Intelligence models \citep{pessach2023algorithmic}. The importance becomes evident when looking at several discriminatory behaviours by algorithms in the past, ranging from hiring \citep{dastin2022amazon} to social welfare systems \citep{hadwick2021lessons}. In recent years, large generative (multimodal) models, particularly Large Language Models (LLMs) with their high accessibility and wide range of applications, have posed significant new challenges to the algorithmic fairness domain \citep{chu2024fairness,kotek2023gender}.

\subsection{Previous literature}
Discrimination through algorithms is not a recent phenomenon and was analysed prior to the AI Act. As early as the 1980s, algorithmic discrimination was observed in admissions settings \citep{connors1981national, williams1981analysis}, and gender inequalities in educational software were discussed \citep{huff1987sex}. In 1996, researchers pointed out the levels at which such technical constraints and social institutions' biases can occur \citep{friedman1996bias}. More recently, the interaction between AI and automated decision-making systems with traditional European and US non-discrimination law has been studied extensively. \citep{barocas2016big} presented a taxonomy of different sources of discrimination and their impact on humans. The relationship among various European non-discrimination laws has also been studied \citep{wachter2021fairness,wachter2020bias,hacker2018teaching, weerts2023algorithmic, xenidis2020eu, lewis2025mapping}. For example, the seminal study by \citet{wachter2021fairness} presented a fairness metric that connects algorithmic fairness to the jurisprudence of the European Court of Justice in relation to the General Data Protection Regulation. Similarly, the connection between non-discrimination law and the GDPR has been analysed to ``unlock the algorithmic black box'' \citep{hacker2018teaching}. Naturally, since these papers were published before the AI Act was passed, they do not contain references to it.
 
Recently, several papers have addressed the AI Act. Short studies such as \citep{deck2024implications, ruohonen2024algorithmic} have analysed the relationship between fairness and the AI Act. Additionally, sections within papers of broader scope beyond pure non-discrimination analyses have mentioned aspects of the relationship between algorithmic fairness and the AI Act \citep{wachter2024limitations,hacker2024generative,novelli2024generative}. \citet{correa2025better} investigated fairness principles in the AI Act in the context of human oversight mechanisms. The paper of \citet{raman2025disclosure} discussed the fairness provisions of the AI Act for General Purpose AI (GPAI) models. The AI Act and its relation to gender equality and non-discrimination law have also been discussed \citep{lutz2024ai}. Among these, the work by \citet{bosoer2023non} is most similar to ours in terms of its focus on the AI Act; however, it investigates non-discrimination provisions in the draft of the AI Act without a strong focus on the interaction with computer science. The most similar work in terms of the interaction between computer science and legal research is the paper by \citet{weerts2023algorithmic}. Unlike our work, theirs did not focus on the AI Act.

\subsection{Our contributions}
This paper aims to foster and extend an interdisciplinary view on explicit and implicit fairness provisions in the AI Act. For the analysis of the EU AI Act's non-discrimination requirements in Sections 2 and 3, our contributions are as follows:
\begin{enumerate}
\item We present the history, scope, and intentions of non-discrimination provisions in the AI Act.
\item We analysed the relation between high-risk systems' provisions and algorithmic fairness, finding that specific provisions will benefit from the forthcoming standardisation process. 
\item We discuss future steps for the application of the AI Act's provisions for algorithmic fairness in the context of classical non-discrimination regulations and the standardisation process.
\end{enumerate}

\section{A primer on EU non-discrimination legislation and algorithmic fairness}
Interdisciplinary research on algorithmic fairness poses challenges for both computer scientists and legal scholars specialising in non-discrimination law. Understanding legal reasoning can be challenging without prior legal knowledge, just as understanding algorithmic methods is difficult without a computational background. 

However, the complexity of interdisciplinary work goes beyond technical expertise. The challenges begin with terminology: For instance, in computer science, ``fairness'' is a term that can refer to different desiderata that aim to prevent socially or morally undesirable behaviour or outcomes of algorithms. Although fairness has been discussed as a principle in economic law contexts \citep{scheuerer2023fairness}, it is not a specific legal term. The closest legal term is arguably ``non-discrimination'', which focuses on preventing unfair treatment based on characteristics such as race, gender, or other attributes on legal grounds. The interaction between these key concepts from law and computer science is demanding.

Therefore, we briefly introduce EU non-discrimination law for computer scientists in the following subsection, and we present an introduction of algorithmic fairness to legal scholars. 

\subsection{EU non-discrimination law: A gentle introduction}
Since the AI Act is influenced by both EU fundamental rights law and traditional EU non-discrimination law, we describe them first.

\textbf{Fundamental rights in EU law.}
The Charter of Fundamental Rights of the European Union (CFR) is today one of the primary fundamental rights laws within the EU\footnote{The CFR should not be confused with the European Convention on Human Rights (ECHR), which, though separate from the EU, is an agreement to which the EU is expected to accede. The relationship between the ECHR and EU law is complex (see \citet{brittain2015relationship}).}. Under Article 6(1) of the Treaty on European Union, the Charter forms part of the primary law of the EU.

The CFR includes non-discrimination law. Article 21 CFR states: ``Any discrimination based on any ground such as sex, race, colour, ethnic or social origin, genetic features, language, religion or belief, political or any other opinion, membership of a national minority, property, birth, disability, age or sexual orientation shall be prohibited.'' Thus, only discrimination based on these listed attributes is targeted.

However, the right to non-discrimination is not absolute. It is important to note that the right to non-discrimination is only one right in the CFR, among others, such as the freedom of expression and information (Article 11(1) CFR). Interpretation and application of the Charter require balancing rights as described in Article 52(1) CFR. First, legislators must balance different CFR rights proportionately when creating new laws. Additionally, when laws reference the CFR, implementers must also follow this proportionality requirement. A logic of proportionality guides the assessment of fundamental rights \citep{almada2023eu}. Fundamental rights are neither absolute, hierarchical, nor quantifiable, and must be applied on a case-by-case basis \citep{sousa2024artificial}. The provisions of the CFR apply to public parties (vertical applicability, \citet{fornasier2015impact}). Therefore, there are debates about how exactly the CFR applies to relationships between private parties (horizontal applicability) in general \citep{fornasier2015impact, cherednychenko2007fundamental,frantziou2015horizontal,prechal2020horizontal} and in the AI Act context specifically \citep{lewis2025mapping}. In the context of the GDPR, scholars and the European Court of Justice (ECJ) noted\footnote{See case C-555/07, Kücükdeveci, para 27.} that the CFR can have a horizontal effect ``when EU secondary law gives expression to a general principle of EU law, such
as the principles of privacy and protection of personal data and non-discrimination.'' \citep{ufert2020ai}. Therefore, some horizontal applicability of the CFR is assumed when applying the AI Act. When discussing fundamental rights in the AI Act, it is also essential to consider the broader political and institutional landscape \citep{palmiotto2025ai}. 

\textbf{Direct and indirect discrimination in EU Law.}
While the CFR is part of EU primary law, it also influences more recent secondary EU law  \citep{lewis2025mapping}. EU law has a long history of non-discrimination regulations\footnote{See for an overview: \url{https://commission.europa.eu/aid-development-cooperation-fundamental-rights/your-fundamental-rights-eu/know-your-rights/equality/non-discrimination_en}.}. For example, the Directive 2000/43/EC (Race Equality Directive) and Directive 2000/78/EC (Employment Equality Directive) include specific non-discrimination norms. 

In order to legally assess discrimination under specific secondary EU law, two types of discrimination are distinguished: direct and indirect discrimination \citep{zuiderveen2024non, wachter2020bias}. However, importantly, in EU law, direct discrimination does not require any intent to discriminate \citep{weerts2023algorithmic,xenidis2020eu, adams2023directly}.

Direct discrimination is defined as situations in which ``one person is treated less favourably than another is, has been or would be treated in a comparable situation'' (Article 2(2)(a) Directive 2000/43/EC)\footnote{In the cited directive, this targets discrimination on grounds of racial or ethnic origin}. This means that discrimination occurs when individuals are treated less favourably based on a protected attribute listed in the Article.

Indirect discrimination is more difficult to address. In indirect discrimination cases, seemingly neutral attributes are used but they rely on a protected attribute. For indirect discrimination, it is important to note that it can be justified through legitimate aims and appropriate means \citep{zuiderveen2024non}.

Also, a third category of discrimination needs to be taken into account\footnote{Please note that other non-discrimination rules in sector-specific regulations, such as the EU Consumer Credit Directive, are beyond the scope of our paper.}, which challenges traditional non-discrimination law: intersectional discrimination. Intersectional discrimination concerns cases ``originating in several inextricably linked vectors of disadvantage'' \citep{xenidis2023computers}. This becomes especially important since (modern) AI models do not use single variables as input but instead use many different aspects as input. This can lead to effects where discrimination may occur only at the intersection of gender and age \citep{weerts2023algorithmic}. Whether, and to what extent, the ECJ currently recognises intersectional discrimination as a distinct form of discrimination remains an open question\citep{weerts2023algorithmic, xenidis2023computers,atrey2018illuminating}.

\subsection{Algorithmic fairness: A gentle introduction}
\label{sec::IntroAlgoritmicFairness}
Algorithmic fairness has emerged as an established computer science and AI-related field in recent years. Algorithmic fairness focuses on uncovering and rectifying disadvantageous treatment of individuals or groups in machine learning models~\citep{mitchell2021algorithmic}.  This treatment must be considered within appropriate social, theoretical, and legal contexts~\citep{pessach2023algorithmic} and includes evaluation and auditing methods to test for unfairness.

\textbf{Origins of unfairness in algorithms.}
Unfair algorithmic behaviour can have different sources. In order to discuss the origins, another --- not well-defined --- term from computer science is often used: bias. There exist many different bias definitions. Since the EU AI Act does not define the word bias itself, we use a definition from the European Commission from 2021. According to this, a bias can be defined as ``[...] bias describes systematic and repeatable errors in a computer system that create unfair outcomes, such as favouring one arbitrary group of users over others'' \citep{noauthor_commission_2021}. Biases are primarily discussed within the data used for AI models and can be sorted into different categories. Most researchers emphasise the different aspects of data inequalities when discussing algorithmic fairness and sort data biases into different categories, such as the over-representation of specific groups in datasets \citep{zou2018ai}. \citep{hacker2018teaching} categorises the biases in data into two groups: biased training data versus unequal ground truth. \citep{barocas2016big} uses five categories to map biases, while \citep{mehrabi2021survey} uses 21 different categories for biases. It is important to note that the data are only one --- albeit important --- source of biases. For example, design decisions in the algorithms themselves or structural power asymmetries also lead to algorithmic unfairness \citep{mehrabi2021survey, gebru2021datasheets, sousa2024artificial}.

\textbf{Approaches to algorithmic fairness: Quantification.}
Based on the different understandings of unfairness and for the purpose of uncovering and rectifying unjustified treatment between groups in algorithms, so-called fairness metrics have been proposed \citep{corbett2023measure,castelnovo2022clarification,verma2018fairness}. The core idea of fairness metrics is to quantify algorithmic outputs and make them comparable through numerical measurements. Most metrics try to assess whether an output of an ML system is unfair to individuals \citep{dwork2012fairness} or groups of people \citep{barocas2023fairness,binns2020apparent}. They provide a quantitative measure intended to indicate whether an algorithm demonstrates unjustified unequal treatment. Fairness metrics also include moral norms \citep{deck2024implications, hellman2020measuring}.

The emergence of the field of algorithmic fairness has challenged existing approaches to ML by highlighting the need to contextualize them, decide on, and provide a rationale for the optimisation criteria chosen. First, it has been shown that an apparently objective number needs to be contextualized to the application under consideration \citep{wachter2021fairness}. Furthermore, some metrics are, from a mathematical viewpoint, incompatible with each other\footnote{They are also incompatible from a moral point of view \citep{heidari2019moral}.}, thus these fairness metrics cannot be satisfied at the same time \citep{chouldechova2017fair,kleinberg2016inherent}. This can lead to an effect of ``d-hacking'' or ``fairness hacking'' where users can choose their favourite metric to create the (technical and mathematical) impression that the algorithm is fair \citep{black2024d,meding2024fairness}.
Fairness metrics for mitigation can play a role at different stages of the development, testing, and deployment of an algorithm. The pre-processing stage, the in-training stage, and the post-processing phases are distinguished \citep{barocas2023fairness, binns2018fairness}. In the pre-processing phase, the input data itself is altered to ensure fair processing \citep{caton2024fairness,kamiran2012data}. For the in-training phase, the optimisation process during the training of an ML model is adjusted to include fairness as an optimisation goal. Finally, in the post-processing phase, the output of a pre-trained model is adapted. The first and last approach make it possible to perform fairness analyses even if one only has black box access to the model \citep{caton2024fairness}.

The era of LLMs raises new questions. LLMs have the advantage --- or disadvantage, depending on the viewpoint --- that their exact use case is most of the time not predefined, and they are trained on large amounts of various data. This diversity makes them later applicable to different contexts, from providing cookie recipes to coding exercises. However, this diversity also includes variations in unfairness. Algorithmic fairness in relation to LLMs is thus somewhat different from algorithmic fairness in classical AI \citep{doan2024fairness,chu2024fairness}. Some researchers have applied classical fairness metrics in classification settings to LLMs \citep{chhikara2024few}. Additionally, it has been shown that LLM-specific issues arise due to the use of massive data and processing \citep{kotek2023gender,ferrara2023should, navigli2023biases, huang2024trustllm} at various steps of algorithmic development. Thus, it was argued that LLMs cannot yield fair outcomes at all \citep{anthis2024impossibility}.

\section{EU AI Act's non-discrimination provisions for high-risk systems.}
Before we discuss the specific non-discrimination obligations for high-risk systems, we will briefly introduce the history, scope, and most important definitions of the AI Act. This short introduction is followed by an analysis of the AI Act's non-discrimination provisions for high-risk systems.

\subsection{The EU AI Act: History, Scope, and Definitions}
In this section, we first describe the origins of the EU AI Act and clarify key concepts such as risk, systems, and the difference between developers and deployers, which is necessary to understand our analysis of the AI Act. 

Emergent technology can benefit from specific regulation. As one of the first comprehensive AI regulations worldwide \citep{wodi2024artificial}, the AI Act (AIA) introduces harmonised rules and provisions for the use and “placing on the market” of AI systems within the European Union (Article 1(2) AIA). The AI Act is a mix of a product safety regulation and fundamental rights protection \citep{almada2023eu}, aiming to establish a level playing field for AI technologies across the Union \citep{nolte2024robustness}.
Notably, the original drafts of the European Commission did not include explicit provisions on individual rights \citep{hacker2024generative}. At that stage, individual rights --- understood as protecting the rights of individuals affected --- were largely absent. In fact, Members of the European Parliament initially did not prioritise the regulation of algorithmic discrimination \citep{chiappetta2023navigating}. However, over the course of the legislative process, different aspects related to fairness were integrated into the final text.

The AI Act relies in part on the New Legislative Framework of the European Union (EU) \citep{kaminski2025american}. The New Legislative Framework is a cornerstone of modern product safety law in the EU (see also \citep{noauthor_commission_2022} for further details), which also applies, for example, to medical devices and children’s toys\footnote{See \url{https://single-market-economy.ec.europa.eu/single-market/goods/new-legislative-framework_en} for an overview.}.

The AI Act regulates AI systems and also includes provisions for General Purpose AI models. The relationship or distinction between a model and a system in the context of the AIA remains legally unresolved. Article 3(1) defines an “AI system” as ``a machine-based system that is designed to operate with varying levels of autonomy and that may exhibit adaptiveness after deployment, and that, for explicit or implicit objectives, infers, from the input it receives, how to generate outputs such as predictions, content, recommendations, or decisions that can influence physical or virtual environments'', but the article does not provide a definition of ``AI model''. Recital 97 clarifies that ``although AI models are essential components of AI systems, they do not constitute AI systems on their own. AI models require the addition of further components, such as for example a user interface, to become AI systems.'' This formulation does not conclusively settle the terminological relationship between models and systems, as discussed in more detail in \citep{nolte2024robustness}. Moreover, the regulation of ``AI systems'' contrasts with the (trustworthy) computer science literature, which typically studies machine learning models. Since our focus is the legal framework, we follow the terminology of the AI Act, even though computer science literature tends to focus on models rather than systems.

The core regulatory structure of the AI Act is built on a risk-based approach~\citep{hacker2024regulating}. Risk is defined in Article 3(2) AIA as “the combination of the probability of the occurrence of harm and the severity of that harm.” Based on this definition, the AI Act categorises AI systems into different risk levels, each associated with a specific set of regulatory requirements\footnote{This classification has been subject to criticism, see e.g., \citep{bosoer2023non}.}:
\begin{itemize}
\item Unacceptable risk (Art. 5 AIA): These AI systems are prohibited. Examples include social scoring by governments.
\item High-risk (Art. 6ff. AIA): These systems are subject to specific provisions, such as requirements for robustness, accuracy, or non-discrimination. Examples include AI used in recruitment, credit scoring, or law enforcement.
\item Certain AI systems (Art. 50 AIA) with specific risks: These systems must meet transparency provisions, such as informing users they are interacting with an AI system. Examples include systems that produce deep-fakes.
\item All other systems: These systems are not subject to specific provisions under the AIA, except broad norms such as Article 4 AIA (AI literacy).
\end{itemize}

In addition, the AIA includes specific provisions for GPAI models (Articles 51 ff. AIA). The use case is not predefined, which is characteristic of GPAI models. LLMs are an example of a typical GPAI model. The AI Act distinguishes between general-purpose models posing ``systemic risk''  and those that do not. 

The primary addressees of the AI Act are providers (Article 3(3) AIA), while secondary addressees are deployers (Article 3(4) AIA). Providers are the entities responsible for developing and placing an AI model on the market or putting an AI system into service. In contrast, deployers are those who use an AI system under their authority. Many of the provisions we discuss in the following sections primarily concern the providers of AI systems, while some apply specifically to deployers (e.g., Article 26 AIA). 

\subsection{Non-discrimination provisions within the AI Act.}

We began by scanning the AI Act for non-discrimination-related terms. In total, we scanned the AI Act for non-discrimination-related terms: discrimination, fundamental right, fairness, and bias. We noticed that within the definitions of Article 3 AIA, the terms serious incident (Article 3(49) AIA) and systemic risk (Article 3(65) AIA) refer to fundamental rights. Thus, these were included in our analysis as well. A full table with all articles of the AI Act, including non-discrimination-related terms, can be found in the appendix, Table \ref{tab:tab2}.

This analysis shows, first, that the majority of non-discrimination provisions in the EU AI Act concern the regulation of high-risk systems. GPAI models are only implicitly regulated by the systemic risk and serious incident terms. This will have an impact on our further analysis: we focus on high-risk systems compared to GPAI models.

\subsection{High-risk AI non-discrimination provisions}
What is a high-risk AI system? Article 6 AIA defines different aspects of high risk. For example, according to Article 6(2) and in combination with Annex III, high-risk systems are, for example, intended to be used as safety components in the management and operation of critical digital infrastructure (Annex III(2) AIA) or systems intended to be used for the recruitment or selection of natural persons (Annex III(4)(a) AIA). The reasoning behind the specific rules in Article 6 AIA is that systems which pose a specific risk to fundamental rights or safety need tighter regulation \citep{fraser2024acceptable} than other, less risky systems. Recital 48 states that the adverse impact caused by the AI system on fundamental rights is of particular relevance when classifying an AI system as a high-risk system. It is estimated that 5\%-15\%\footnote{For a critical assessment of these numbers see \citep{almada2023eu}} of all AI systems fall under the category of high-risk systems \citep{noauthor_commission_2021}.

We identified Articles 9, 10, and 15 AIA as the main non-discrimination provisions of high-risk systems. These will be discussed in detail in the following paragraphs\footnote{Please note that we do not distinguish here between the provisions for providers and those provisions for the deployers of AI systems, since we focus on the interaction of computer science and law.}. Furthermore, Article 11 (together with Annex IV) and Article 13 AIA will be discussed due to their relationship to the aforementioned articles. 

\textbf{Article 9 AIA: Non-discrimination provisions within risk management systems}
Article 9 of the AI Act covers risk management systems of high-risk systems. According to Article 9(2)(a-d) AIA, risk management systems are designed to identify, evaluate, and mitigate risks of AI systems in a continuous, iterative process throughout the entire lifecycle of a high-risk AI system. All risks after mitigation, called residual risks, need to be judged acceptable \citep{soler2023analysis}.

Of particular relevance to non-discrimination law is Article 9(2)(a) AIA, which requires the identification and analysis of known and reasonably foreseeable risks\footnote{See for the definition of a foreseeable risk also the EU Blue Guide of the New Legislative Framework \citet{noauthor_commission_2022}.} that the high-risk AI system can pose to health, safety, or fundamental rights. Through the link to fundamental rights, indirectly, Article 9 AIA already requires the identification and analysis of non-discrimination issues \citep{zuiderveen2024non}. The initial draft of the AI Act did not include the reference to fundamental rights, and it was added by the EU Council in 2022\footnote{See Council of the European Union, ``Proposal for a Regulation of the European Parliament and the Council laying down harmonised rules on artificial intelligence (Artificial Intelligence Act) and amending certain Union legislative acts – General approach'' Doc. 14954/22, 25 November 2022.}, highlighting the legislator's intent to include non-discrimination aspects in the AI Act.

\textbf{Article 10(2)(f) AIA: Main input non-discrimination provision of high-risk systems.}
The core of the non-discrimination provision for high-risk systems is Article  10(2)(f) AIA. Article 10(2)(f) AIA requires that training, validation, and testing datasets shall be subject to an ``examination in view of possible biases that are likely to affect the health and safety of persons, have a negative impact on fundamental rights or lead to discrimination prohibited under Union law, especially where data outputs influence inputs for future operations''.

The first observation we make is that the input data of an AI system should undergo a bias analysis. Most importantly, the term \textit{bias} is neither defined in the AI Act nor is there a common understanding of its meaning \citep{van2024using}. It seems that the regulators had a more technical definition of bias in mind, focusing on the diversity of training data in different dimensions compared to social, ethical, or structural biases \citep{hacker2024generative}. This could imply difficulties in determining the regulatory content, also for the later standardisation process. 

The wording of Article 10(2)(f) AIA names biases that impact fundamental rights or lead to discrimination prohibited under Union law. Thus, on the one hand, there is at least some link to EU non-discrimination law. There is a possibility that various aspects of discriminatory effects are covered. One can argue that the link achieves a strong protection objective for the input side of the AI systems. However, even without this link, Union law, which prohibits discrimination, could be applicable (see also next section).

Article 10(2)(f) AIA only targets the input data for machine learning. Under a strict interpretation, the examination may only be mandatory for training, validation, and testing data. In other words, it only covers the input data of an AI system. Although the second half of the sentence reads ``especially where data outputs influence inputs for future operations'', output data itself is not the main focus. 

Furthermore, the wording of Article 10(2)(f) AIA could imply an expost view: the provider has to know whether a bias will likely lead to discrimination. However, especially for larger models, it is technically very challenging, if not impossible, to predict the individual output of AI models \citep{hacker2024generative, black2022model}.

\textbf{Article 10(2)(g) AIA: Appropriate bias detection, prevention, and mitigation on input data must include factors beyond technical means.}
Article 10(2)(f) AIA is complemented by Article 10(2)(g) AIA. Article 10(2)(g) AIA demands that ``appropriate measures to detect, prevent and mitigate possible biases identified according to point (f)'' must be implemented. Thus, Article 10 AIA not only mandates the examination of biases that lead to discrimination but also the mitigation of these biases. The bias tests must be documented and disclosed according to Article 11 AIA, along with Annexes IV and IXa AIA \citep{zuiderveen2024non}. It has been argued that with this focus on the input side, the AI Act seeks to remedy the root cause of biases \citep{zuiderveen2024non}. 

Article 10(2)(g) AIA appears to be inspired by the technical literature on fairness metrics for algorithmic outputs. Nevertheless, Article 10(2)(g) AIA targets the input of algorithms. Therefore, classical fairness metrics for algorithmic outputs do not apply directly. Methods for detecting, preventing, and mitigating input bias remain relatively underexplored in the existing literature, and these methods are not clearly defined in the AI Act \citep{wachter2024limitations, deck2024mapping}. 

\textbf{Article 10(2) AIA: Additional provisions without specific algorithmic fairness implications.}
In reviewing Article 10(2) AIA, it is noticeable that the other sections of Article 10(2) AIA also indirectly link to algorithmic fairness considerations. For example, the representativeness of training data is explicitly mentioned in Article 10(2)(d). When interpreting Article 10(2)(f) AIA or Article 10(2)(g) AIA, however, the additional requirements in Article 10(2) AIA might not lead to different results. Since these articles have no direct requirements regarding non-discrimination, we will not analyse them in further detail.

\textbf{Article 10(5) AIA: A legal basis for the processing of personal data in non-discrimination contexts.}
Finally, Article 10(5) AIA allows the processing of Article 9 GDPR data (see also Recital 70 AIA). Article 9 GDPR protects special categories of personal data, such as genetic, biometric, or health data (Article 9(1) GDPR). There is a tension \citep{deck2024implications} between the need for debiasing AI algorithms and data protection law, which Article 10(5) AIA aims to address. In order to effectively mitigate biases in AI systems, the processing of personal data (for example, to compute fairness metrics) is important \citep{van2024using}. Notably, this exception only applies in the high-risk regime. Thus, this exception is not applicable to non-high-risk systems. Nevertheless, Article 10(5) is an example of how the AI Act not only imposes burdens on providers of AI systems but also grants rights to these groups. This aspect is frequently underappreciated in the discourse regarding the AI Act.

\textbf{Article 11(1) AIA and Annex IV(2)(g) AIA: Technical documentation of bias testing methods.}
Article 11(1) together with Annex IV(2)(g) AIA requires that the technical documentation of a high-risk system includes ``information about the validation and testing data used and their main characteristics; metrics used to measure accuracy, robustness and compliance with other relevant requirements set out in Chapter III, Section 2, as well as potentially discriminatory impacts''. 

\textbf{Article 13 AIA: Transparency and provision of information to deployers.}
Article 13(1) AIA provides that `High-risk AI systems shall be designed and developed in such a way as to ensure that their operation is sufficiently transparent to enable deployers to interpret a system’s output and use it appropriately''.  One could contend that the interpretation of a system’s output necessarily entails an assessment of potential discrimination. Recital 72 AIA enumerates aspects regarding the information that should be provided, with specific reference to the protection of fundamental rights. However, the risk identified in the recital is presented in general terms and fails to specify any particular methodology or technological tool that must be employed. Moreover, as Recitals serve a non-binding role and do not possess direct normative force in the interpretation of Article 13 AIA, the lack of specificity in Recital 72 AIA does not alter the conclusion. Accordingly, Article 13(1) AIA itself does not likely impose any obligation to use algorithmic fairness measures or metrics derived from the computer science domain.

Furthermore, even assuming, arguendo, that the use of such algorithmic fairness tools is required during the development or deployment process, Article 13(1) AIA merely obliges providers to document the results. The imposition of mitigation or prevention measures does not fall within the mandatory scope of Article 13(1) AIA.

\textbf{Article 15 (4) AIA: Output bias-related provisions in cases of feedback loops.}
In contrast to Article 10(2)(f) AIA, Article 15(4) AIA mandates that ``High-risk AI systems that continue to learn after being placed on the market or put into service shall be developed in such a way as to eliminate or reduce as far as possible the risk of possibly biased outputs influencing input for future operations (feedback loops), and as to ensure that any such feedback loops are duly addressed with appropriate mitigation measures.'' In this case, the output of a system is regulated, in contrast to the input of an AI system in Article 10 AIA. The reasoning behind this is that AI systems should not become echo chambers for biases \citep{zuiderveen2024non}. According to Recital 67 AIA, this is a particular concern when examining historical biases. It is important to note that this output provision only applies if the systems continue to learn after being placed on the market or put into service.

\textbf{Other articles regarding high-risk systems requirements.}
While the previously discussed articles provide at least some intuition about the addressed concerns, we categorise the remaining articles of Table \ref{tab:tab2}. Most of these paragraphs only contain links to fundamental rights, which show a limited impact on non-discrimination requirements. Thus, the discussion here is rather short.
Article 13(3)(b)(iii) AIA mandates that information on the foreseeable misuse and the impact on fundamental rights be provided. Article 14(2) AIA mandates that human oversight is needed to minimise risks to fundamental rights.
Article 17(1)(i) AIA and Article 26(5) AIA are related to reporting and information on serious incidents. Serious incidents are linked to discrimination through Article 3(49)(c) AIA: ``serious incident'' means an incident or malfunctioning of an AI system that directly or indirectly leads to any of the following: [...] the infringement of obligations under Union law intended to protect fundamental rights''.  Finally, Article 27(1) AIA requires a fundamental rights impact assessment when the high-risk system is used in specific systems within bodies governed by public law. Yet again, significant gaps exist between theoretical and methodological elaboration of these impact assessments \citep{mantelero2024fundamental}. Article 27(1) AIA only addresses public parties. A fundamental rights impact assessment is not necessary for private parties.

Finally, the impact of fundamental rights on non-discrimination is also monitored by a specific authority established through Article 77 (1) and Article 77 (2) AIA. The effect and impact of this authority framework should be assessed and evaluated once it is established. 

\textbf{Summary: High-risk systems balancing between specificity and generality.}
The AI Act aims to be a specific law for AI \citep{zHodi2024conflict}. For the ML and Law community, it might be challenging to understand the implications of this specific AI law. Some articles mention non-discrimination directly, while others focus on fundamental rights. Only Article 10 AIA and Article 15 AIA explicitly mention non-discrimination and biases, compared to the statement of fundamental rights protection in Article 9 AIA. However, Article 10 AIA only considers the input phase, while Article 15 AIA only addresses the output phase for feedback loops. It is unclear why the AI Act does not consistently regulate direct, indirect, and intersectional discrimination across both the input and output phases.

Furthermore, it is essential to note that most provisions predominantly focus on the providers of AI systems from the perspective of a product safety regulation. However, providers may not always be aware of the specific use cases in which a high-risk system will be deployed \citep{correa2025better}. Consequently, fairness testing conducted solely by the provider, regardless of the system development phase, may be insufficiently comprehensive or contextually inappropriate.

\section{The role of standardisation in the AI Act}
According to the New Legislative Framework, all AI systems and models entering the market require a conformity assessment demonstrating that AI systems and models comply with the provisions of the AI Act \citep{kaminski2025american}. This process will culminate in the establishment of harmonised technical standards \citep{ebers2021standardizing}.

The details of the conformity assessments are set out in Articles 40 ff. AIA\footnote{For the process of establishing a harmonised standard see \citep{kaminski2025american}. }. Depending on the specific product, either a self-assessment or a third-party assessment is possible (Article 43 AIA). The AI Act stipulates the requirements for the conformity assessments themselves but does not prescribe the specific content of these assessments \citep{kaminski2025american}.

Under Articles 40 and following, mandates have been issued to the European standardisation organisations. In March 2023, the European Commission issued Mandate M/593 to CEN (European Committee for Standardisation) and CENELEC JTC 21 (European Committee for Electrotechnical Standardisation) \citep{kilian2025european} (updated to Mandate M/613 in 2025). Thus, the details of non-discrimination procedures --- especially concerning methods and the entire risk management framework outlined in Article 17 --- will be developed through the standardisation process. A similar approach exists for GPAI models. For GPAI models, providers could follow the developed Code of Practice to show compliance with the AI Act (Article 55(2) AIA). Throughout the standardisation, experts and stakeholders will collaboratively refine non-discrimination measures for high-risk systems.

However, although standardisation committees have considerable discretion, they must operate within the bounds of legal norms. For example, if the AI Act demands discrimination testing for input data (and excludes output data), standardisation cannot necessarily go beyond this, though it may at times do so. Thus, standardisation in the context of algorithmic fairness can also produce a false sense of safety \citep{laux2024three}.

\section{Interplay between traditional EU non-discrimination law and the AI Act for high-risk systems}
With the adoption of the AI Act, questions arise as to how its regime will interact with pre-existing non-discrimination law in the EU. Whereas the AI Act primarily regulates the input side for high-risk systems, much of the machine learning research community’s work on fairness concerns algorithmic outputs. The AI Act could be regarded as a lex specialis \citep{craig2011eu} so it prevails over general frameworks to the extent that it covers an issue; where it does not address certain discriminatory outcomes, however, the more general non-discrimination law remains applicable. Since the AI Act does not directly regulate algorithmic outputs for standard high-risk systems, traditional EU non-discrimination law may still apply in these cases.

The relationship between the AI Act and existing fundamental rights protections, including those concerning non-discrimination, requires further clarification and research \citep{lewis2025mapping}. This also includes an analysis of when fundamental rights are vertically and horizontally applicable. Furthermore, the fundamental rights in the AI Act need to be balanced \citep{kusche2024possible}, and the material scope is limited due to the enumeration of protected attributes \citep{xenidis2020eu}. As pointed out, ``gaps in the regulatory framework have left fundamental rights inconsistently protected in the AI Act.'' \citep{palmiotto2025ai} Nevertheless, since the AI Act does not provide protection for individuals regarding the output of algorithms, there is room for the applicability of the CFR. It remains an open research question how the protection in the CFR and the AI Act relate \citep{lewis2025mapping}.

Secondary EU law, such as Directives 2004/113/EC and 2000/43/EC, provides more concrete protections against discrimination, and these instruments have clear horizontal applicability, unlike the frequently debated scope of the CFR \citep{xenidis2020eu,lewis2025mapping}. For example, Directives 2004/113/EC (Equal Treatment in Goods and Services Directive) and 2000/43/EC (Race and Ethnicity Equality Directive) specifically target non-discrimination regulations. These regulations address specific cases in which discrimination can occur. Due to their specificity, they have a limited material and personal scope as a result of the enumeration of protected attributes \citep{xenidis2020eu, circiumaru2024discrimination}. The development of the relationship between existing regulations and the AI Act, and the stance that national and EU courts will ultimately adopt, is still unresolved. In the past, the ECJ has already issued important rulings on how the non-discrimination provisions in Article 21 CFR must be interpreted in relation to secondary EU law \citep{xenidis2020eu}. It is very likely that in the future, the ECJ will continue to issue rulings to clarify the interplay between the EU AI Act, fundamental rights, and other directives and regulations. 

The AI Act holds the potential to complement traditional non-discrimination legislation. Traditional non-discrimination law primarily targets the protection of individuals. The EU AI Act and its rules on high-risk systems are fundamentally aligned with product safety regulation rather than human rights protection. Despite the identified challenges associated with this product-safety approach of the AI Act, it nonetheless provides a degree of protection. However, such an emphasis on products can overlook broader regulatory dimensions. In the context of EU data protection, it has been posited that a dual approach involving both individual protection and collaborative governance is necessary \citep{kaminski2018binary}. A dual approach might also be beneficial for non-discrimination law in the context of AI.
\vspace{-0.2cm}
\section{Summary \& Outlook}
As \citet{mayson2018bias} aptly pointed out: ``Bias in, bias out.'' The AI Act encounters significant challenges in regulating such biases. The non-discrimination requirements analysed in the previous section primarily focus on bias in (training) data and, from our perspective, tend to overlook other sources of bias, such as design choices within the algorithms themselves. Many of the rules emphasise (internal) compliance \citep{zuiderveen2024non} and rely on checklists, rather than offering output-based safeguards for users affected by AI systems. This focus stems from the AI Act’s foundations in the New Legislative Framework for product safety law, which, unlike the GDPR, does not incorporate a clear individual rights dimension. Whether self-assessment suffices to protect individuals’ fundamental rights and fulfil non-discrimination obligations remains an open question \citep{bosoer2023non}. This is an area that will require further research, particularly once harmonised standards have been published.

Nevertheless, classical EU non-discrimination law will continue to play an important role in the context of AI. Further research is required to determine how the AI Act’s high-risk systems product safety-based provisions can be reconciled with classical, individual-focused non-discrimination law. In the introduction to our work, we raised the question of how fair outcomes from AI algorithms can be ensured. While the AI Act provides some direction, significant challenges persist at the intersection of algorithmic fairness and non-discrimination law. Our paper seeks to bridge the gap between legal and computational disciplines, which, in our view, will benefit from closer interdisciplinary collaboration.

\section{Acknowledgments}
We would like to thank Miriam Rateike and Michèle Finck for their detailed comments on the manuscript. We are grateful for comments from Christoph Kern and the Social Data Science and Statistical Learning group during an institute seminar, which clearly improved this paper. Furthermore, we thank Alina Wenick, Tharushi Abeynayaka, and Mara Seyfert for their comments on the manuscript and for proofreading the paper.

Kristof Meding was supported by the Carl Zeiss Foundation through the CZS Institute for AI and Law. Additionally, Kristof Meding is a member of the Machine Learning Cluster of Excellence, funded by the Deutsche Forschungsgemeinschaft (DFG, German Research Foundation) under Germany’s Excellence Strategy – EXC number 2064/1 – Project number 390727645.

\bibliography{sample-base}
\bibliographystyle{plainnat}

\appendix

\subsection{Table algorithmic fairness related articles}

\begin{table}[h!]
\vspace{-0.5cm}
\rotatebox{90}{
\tiny
\begin{tabular}{|l|l|l|l|l|l|l|l|l|}
\hline
\multicolumn{1}{|c|}{Article} &
  \multicolumn{1}{c|}{Discrimination} &
  \multicolumn{1}{c|}{\begin{tabular}[c]{@{}c@{}}Fundamental\\  Rights\end{tabular}} &
  \multicolumn{1}{c|}{Fair {[}...{]}} &
  \multicolumn{1}{c|}{Bias} &
  \multicolumn{1}{c|}{\begin{tabular}[c]{@{}c@{}}Systemic\\ risk\end{tabular}} &
  \multicolumn{1}{c|}{\begin{tabular}[c]{@{}c@{}}Serious\\ incident\end{tabular}} &
  \multicolumn{1}{c|}{Specific Topic} &
  \multicolumn{1}{c|}{General Topic} \\ \hline
1 &
   &
  x &
   &
   &
   &
   &
  Subject matter of AI Act &
  \multirow{3}{*}{General Provision} \\ \cline{1-8}
2 &
   &
  x &
   &
   &
   &
   &
  Scope of the AI Act &
   \\ \cline{1-8}
3 &
   &
  x &
   &
   &
  x &
   &
  Definitions &
   \\ \hline
6 &
   &
  x &
   &
   &
   &
   &
  Classification rules for high-risk AI systems &
  \multirow{10}{*}{rovisions for High-Risk Systems} \\ \cline{1-8}
7 &
   &
  x &
   &
   &
   &
   &
  Amendments to Annex III AI Act &
   \\ \cline{1-8}
9 &
   &
  x &
   &
   &
   &
   &
  Risk management system for High-Risk AI Systems &
   \\ \cline{1-8}
10 &
  x &
  x &
   &
  x &
   &
   &
  Data and data governance for High-Risk AI Systems &
   \\ \cline{1-8}
13 &
   &
  x &
   &
   &
   &
   &
  Transparency and provision of information to deployers for High-Risk AI Systems &
   \\ \cline{1-8}
14 &
   &
  x &
   &
  x &
   &
   &
  Human oversight of High-risk Systems &
   \\ \cline{1-8}
15 &
   &
   &
   &
  x &
   &
   &
  Accuracy, robustness and cybersecurity &
   \\ \cline{1-8}
17 &
   &
   &
   &
   &
   &
  x &
  Quality management system &
   \\ \cline{1-8}
26 &
   &
   &
   &
   &
   &
  x &
  Obligations of deployers of high-risk AI systems &
   \\ \cline{1-8}
27 &
   &
  x &
   &
   &
   &
   &
  Fundamental rights impact assessment for high-risk AI systems &
   \\ \hline
28 &
   &
  x &
   &
   &
   &
   &
  Notifying authorities &
  \multirow{2}{*}{Notifying authorities and notified bodies} \\ \cline{1-8}
36 &
   &
  x &
   &
   &
   &
   &
  Changes to notifications of the notifying authority &
   \\ \hline
40 &
   &
  x &
   &
   &
   &
   &
  Harmonised standards and standardisation deliverables &
  \multirow{3}{*}{\begin{tabular}[c]{@{}l@{}}Standards, conformity assessment, \\ certificates, registration\end{tabular}} \\ \cline{1-8}
41 &
   &
  x &
   &
   &
   &
   &
  Common specifications &
   \\ \cline{1-8}
43 &
   &
  x &
   &
   &
   &
   &
  Conformity assessment &
   \\ \hline
51 &
   &
   &
   &
   &
  x &
   &
  Classification of GPAI models as general-purpose AI models with systemic risk &
  \multirow{6}{*}{General Purpose AI} \\ \cline{1-8}
52 &
   &
   &
   &
   &
  x &
   &
  Notification Procedure for GPAI models &
   \\ \cline{1-8}
53 &
   &
   &
   &
   &
  x &
   &
  Obligations for providers of general-purpose AI models &
   \\ \cline{1-8}
54 &
   &
   &
   &
   &
  x &
   &
  Authorised representatives of providers of general-purpose AI models &
   \\ \cline{1-8}
55 &
   &
   &
   &
   &
  x &
  x &
  Obligations of providers of general-purpose AI models with systemic risk &
   \\ \cline{1-8}
56 &
   &
   &
   &
   &
  x &
   &
  Codes of practice for GPAI Models &
   \\ \hline
57 &
   &
  x &
   &
   &
   &
   &
  AI regulatory sandboxes &
  \multirow{3}{*}{Meassures to support innovation} \\ \cline{1-8}
58 &
   &
  x &
  x &
   &
   &
   &
  Detailed arrangements for, and functioning of, AI regulatory sandboxes &
   \\ \cline{1-8}
60 &
   &
   &
   &
   &
   &
  x &
  Testing of high-risk AI systems in real world conditions outside AI regulatory sandboxes &
   \\ \hline
66 &
   &
  x &
   &
   &
   &
  x &
  Tasks of the  European Artificial Intelligence Board &
  \multirow{4}{*}{Governance} \\ \cline{1-8}
67 &
   &
  x &
   &
   &
   &
   &
  Advisory forum &
   \\ \cline{1-8}
68 &
   &
   &
  x &
   &
  x &
   &
  Scientific panel of independent experts &
   \\ \cline{1-8}
70 &
   &
  x &
   &
  x &
   &
   &
  Designation of national competent authorities and single points of contact &
   \\ \hline
73 &
   &
   &
   &
   &
   &
  x &
  Reporting of serious incidents &
  \multirow{9}{*}{\begin{tabular}[c]{@{}l@{}}Post-market monitoring, information sharing\\  and market surveillance\end{tabular}} \\ \cline{1-8}
76 &
   &
   &
   &
   &
   &
  x &
  Supervision of testing in real world conditions by market surveillance authorities &
   \\ \cline{1-8}
77 &
  x &
  x &
   &
   &
   &
   &
  Powers of authorities protecting fundamental rights &
   \\ \cline{1-8}
79 &
   &
  x &
   &
   &
   &
   &
  Procedure at national level for dealing with AI systems presenting a risk &
   \\ \cline{1-8}
82 &
   &
  x &
   &
   &
   &
   &
  Compliant AI systems which present a risk &
   \\ \cline{1-8}
86 &
   &
  x &
   &
   &
   &
   &
  Right to explanation of individual decision-making &
   \\ \cline{1-8}
90 &
   &
   &
   &
   &
  x &
   &
  Alerts of systemic risks by the scientific panel &
   \\ \cline{1-8}
92 &
   &
   &
   &
   &
  x &
   &
  Power to conduct evaluations &
   \\ \cline{1-8}
93 &
   &
   &
   &
   &
  x &
   &
  Power to request measures &
   \\ \hline
101 &
   &
   &
   &
   &
  x &
   &
  Fines for providers of general-purpose AI models &
  Penalties \\ \hline
112 &
   &
  x &
   &
   &
   &
   &
  Evaluation and review Article 112 of the AI Act &
  Final Provisions \\ \hline

\end{tabular}
}
\caption{All articles in the EU AI Act mentioning fairness-related terms.}
  \label{tab:tab2}
\end{table}

\newpage
\section*{NeurIPS Paper Checklist}

\begin{enumerate}

\item {\bf Claims}
    \item[] Question: Do the main claims made in the abstract and introduction accurately reflect the paper's contributions and scope?
    \item[] Answer: \answerYes{} %
    \item[] Justification:  We provide an introduction to the AI Act and analyse non-discrimnation regulations for high-risk systems.

\item {\bf Limitations}
    \item[] Question: Does the paper discuss the limitations of the work performed by the authors?
    \item[] Answer: \answerYes{} %
    \item[] Justification:   We only analyse non-discrimnation regulations for high-risk systems.

\item {\bf Theory assumptions and proofs}
    \item[] Question: For each theoretical result, does the paper provide the full set of assumptions and a complete (and correct) proof?
    \item[] Answer: \answerNA{} %
    \item[] Justification: 

    \item {\bf Experimental result reproducibility}
    \item[] Question: Does the paper fully disclose all the information needed to reproduce the main experimental results of the paper to the extent that it affects the main claims and/or conclusions of the paper (regardless of whether the code and data are provided or not)?
    \item[] Answer: \answerNA{} %
    \item[] Justification: 

\item {\bf Open access to data and code}
    \item[] Question: Does the paper provide open access to the data and code, with sufficient instructions to faithfully reproduce the main experimental results, as described in supplemental material?
    \item[] Answer: \answerNA{} %
    \item[] Justification: 

\item {\bf Experimental setting/details}
    \item[] Question: Does the paper specify all the training and test details (e.g., data splits, hyperparameters, how they were chosen, type of optimizer, etc.) necessary to understand the results?
    \item[] Answer: \answerNA{} %
    \item[] Justification:

\item {\bf Experiment statistical significance}
    \item[] Question: Does the paper report error bars suitably and correctly defined or other appropriate information about the statistical significance of the experiments?
    \item[] Answer: \answerNA{}%
    \item[] Justification: 

\item {\bf Experiments compute resources}
    \item[] Question: For each experiment, does the paper provide sufficient information on the computer resources (type of compute workers, memory, time of execution) needed to reproduce the experiments?
    \item[] Answer: %
    \item[] Justification: \answerNA{}
    
\item {\bf Code of ethics}
    \item[] Question: Does the research conducted in the paper conform, in every respect, with the NeurIPS Code of Ethics \url{https://neurips.cc/public/EthicsGuidelines}?
    \item[] Answer:  answerYes{}%
    \item[] Justification: 

\item {\bf Broader impacts}
    \item[] Question: Does the paper discuss both potential positive societal impacts and negative societal impacts of the work performed?
    \item[] Answer: \answerYes{}, %
    \item[] Justification: We focus on non-discrimination regulations in the EU AI Act.
    
\item {\bf Safeguards}
    \item[] Question: Does the paper describe safeguards that have been put in place for responsible release of data or models that have a high risk for misuse (e.g., pretrained language models, image generators, or scraped datasets)?
    \item[] Answer: \answerNA{} %
    \item[] Justification:

\item {\bf Licenses for existing assets}
    \item[] Question: Are the creators or original owners of assets (e.g., code, data, models), used in the paper, properly credited and are the license and terms of use explicitly mentioned and properly respected?
    \item[] Answer: \answerNA{} %
    \item[] Justification: 

\item {\bf New assets}
    \item[] Question: Are new assets introduced in the paper well documented and is the documentation provided alongside the assets?
    \item[] Answer: \answerNA{} %
    \item[] Justification: 

\item {\bf Crowdsourcing and research with human subjects}
    \item[] Question: For crowdsourcing experiments and research with human subjects, does the paper include the full text of instructions given to participants and screenshots, if applicable, as well as details about compensation (if any)? 
    \item[] Answer: \answerNA{}. %
    \item[] Justification: 

\item {\bf Institutional review board (IRB) approvals or equivalent for research with human subjects}
    \item[] Question: Does the paper describe potential risks incurred by study participants, whether such risks were disclosed to the subjects, and whether Institutional Review Board (IRB) approvals (or an equivalent approval/review based on the requirements of your country or institution) were obtained?
    \item[] Answer: \answerNA{}%
    \item[] Justification: 

\item {\bf Declaration of LLM usage}
    \item[] Question: Does the paper describe the usage of LLMs if it is an important, original, or non-standard component of the core methods in this research? Note that if the LLM is used only for writing, editing, or formatting purposes and does not impact the core methodology, scientific rigorousness, or originality of the research, declaration is not required.
    \item[] Answer: \answerNA{} %
    \item[] Justification: 

\end{enumerate}

\end{document}